\documentclass[conference]{IEEEtran}
\IEEEoverridecommandlockouts

\usepackage{cite}
\usepackage{amsmath,amssymb,amsfonts}
\usepackage{amsbsy}
\usepackage{graphicx}
\usepackage{textcomp}
\usepackage{xcolor}
\usepackage{mwe}

\ifCLASSOPTIONcompsoc
  \usepackage[caption=false,font=normalsize,labelfont=sf,textfont=sf]{subfig}
\else
  \usepackage[caption=false,font=footnotesize]{subfig}
\fi
\usepackage{algorithm}
\usepackage[noend]{algpseudocode}
\usepackage{comment}
\usepackage{theorem}
\usepackage{amsmath,xparse}
\usepackage{lastpage}
\usepackage{times}  
\usepackage{helvet}  
\usepackage{courier}  
\usepackage{url}  
\usepackage{xcolor}
\usepackage{tikz}
\usepackage{pgflibraryarrows}
\usepackage{pgflibrarysnakes}
\usetikzlibrary{arrows,shapes}
\usepackage[mathscr]{eucal}
\usetikzlibrary{positioning,calc}
\usepackage{float}
\usepackage{stmaryrd}
\def\BibTeX{{\rm B\kern-.05em{\sc i\kern-.025em b}\kern-.08em
    T\kern-.1667em\lower.7ex\hbox{E}\kern-.125emX}}
\begin{document}

\title{Tensor-Train Parameterization for Ultra Dimensionality Reduction}

\author{\IEEEauthorblockN{1\textsuperscript{st} Mingyuan Bai}
\IEEEauthorblockA{\textit{Discipline of Business Analytics} \\
\textit{The University of Sydney Business School}\\
\textit{The University of Sydney}\\
Camperdown, NSW, Australia \\
mbai8854@uni.sydney.edu.au}
\and
\IEEEauthorblockN{2\textsuperscript{nd}  S.T. Boris Choy}
\IEEEauthorblockA{\textit{Discipline of Business Analytics} \\
\textit{The University of Sydney Business School}\\
\textit{The University of Sydney}\\
Camperdown, NSW, Australia \\
boris.choy@sydney.edu.au}
\and
\IEEEauthorblockN{3\textsuperscript{rd} Xin Song}
\IEEEauthorblockA{\textit{Discipline of Business Analytics} \\
\textit{The University of Sydney Business School}\\
\textit{The University of Sydney}\\
Camperdown, NSW, Australia \\
\textit{School of Computer Science} \\
\textit{China University of Geosciences}\\
Wuhan 430074, P. R. China \\
xson3983@uni.sydney.edu.au}
\and
\IEEEauthorblockN{4\textsuperscript{th} Junbin Gao}
\IEEEauthorblockA{\textit{Discipline of Business Analytics} \\
\textit{The University of Sydney Business School}\\
\textit{The University of Sydney}\\
Camperdown, NSW, Australia \\
junbin.gao@sydney.edu.au}
}

\maketitle

\begin{abstract}
Locality preserving projections (LPP) are a classical dimensionality reduction method based on data graph information. However, LPP is still responsive to extreme outliers. LPP aiming for vectorial data may undermine data structural information when it is applied to multidimensional data. Besides, it assumes the dimension of data to be smaller than the number of instances, which is not suitable for high-dimensional data. For high-dimensional data analysis, 
the tensor-train decomposition is proved to be able to efficiently and effectively capture the spatial relations.
Thus, we propose a tensor-train parameterization for ultra dimensionality reduction (TTPUDR) in which the traditional LPP mapping is tensorized in terms of tensor-trains and the LPP objective is replaced with the Frobenius norm to increase the robustness of the model. The manifold optimization technique is utilized to solve the new model. The performance of TTPUDR is assessed on classification problems and TTPUDR significantly outperforms the past methods and the several state-of-the-art methods. 
\end{abstract}


\section{Introduction}\label{Sec:1}
The ultra high-dimensional data, attracting great attention from both academia and the industry, have been common in computer vision \cite{TensorFace}, recommender systems \cite{TensorRecomd}, signal processing \cite{TensorSignal} and neuroscience \cite{TensorFMRI}. In many cases, high-dimensional data are converted from the so-called multi-dimensional data, commonly referred to as tensors or multi-arrays. 
There exists a great amount of research which scrutinizes the information in the tensors. To avoid the curse-of-dimensionality issue in data-driven learning, research on dimensionality reduction by taking the tensorial structure into account has attracted great interests in literature \cite{WangAggarwalAeron2016,TLPP,WangAggarwalAeron2018}. 

Many methods are utilized to explore the information in tensors
by tensor decomposition methods. A group of them presume and maintain
spatial structures in tensors.  
Three classical methods are the CANDECOMP/PARAFAC (CP) decomposition \cite{CPDecomp}, the Tucker decomposition \cite{TuckerDecomp} and tensor-train (TT) decomposition \cite{TTDecomp}. 
The TT decomposition offers the most compact capacity by decomposing an $n$-order tensor 
in terms of the multiplication of $n$ 3-order core tensors in a chain.
The TT decomposition has comparatively lower storage complexity against others with acceptable accuracy. 
Given its capacity, the TT decomposition can avoid the curse of dimensionality and thus is more appropriate to the analysis of higher-mode tensors or ultra-dimensional vectors. 

Although the tensor decomposition methods, especially the TT decomposition, are relatively efficient to form a tensor subspace with sufficient spatial relational information for high-mode tensors, it should be noted that the computational and storage cost of including redundant information in the data is also an issue.
A number of dimensionality and feature extraction methods have been proposed and implemented in the past decades.
Two of the most powerful, renowned and classical ones are the principal component analysis (PCA) \cite{PCA} and the locality preserving projections (LPP) \cite{LPP}. 
Yet PCA is significantly sensitive to the outliers and focuses more on the global information. LPP, on the contrary, is concerned with the local information of the data and has a lower sensitivity to the outliers than PCA by minimizing the squared Frobenius norm of the distance between the data in the lower-dimensional space. Nevertheless, LPP is still evidently affected by extreme outliers. Therefore, a more robust dimensionality reduction or feature extraction method should be implemented. 
It is well known that the $\ell_1$-norm is robust to outliers, including the extreme ones, which has been applied in both PCA and LPP. 
However, the $\ell_1$-norm is not differentiable at every point.
Furthermore, minimizing an $\ell_1$-norm objective function with respect to a matrix optimization variable is in substance minimizing each element or column vector of this matrix variable, where these elements or column vectors are the components of this matrix variable. Thus, the spatial relation is not considered sufficiently. An approximation to the $\ell_1$-norm is the Frobenius norm. When minimizing the Frobenius norm objective function, all the components are treated as a whole group and the spatial relations between the components are thus adequately preserved and analyzed.
Most of the existing dimensionality reduction methods are applied to high/multi-dimensional data by vectorizing them. This vectorization enlarges the parameter space of the algorithms and neglects the spatial relational information existing in multi-way data. 
Therefore, the tensor subspace embedded dimensionality reduction methods come on stage. There are already a small number of existing attempts to embed the tensor subspace into the low-dimensional spaces. 
For example, 
Tucker LPP (TLPP) \cite{TLPP} embeds the tensor subspace based on the Tucker decomposition into the low-dimensional space under the LPP criterion. The local relation is sufficiently captured, but the accuracy is deteriorated due to the sensitivity to the exceptional outliers and the computational cost also exponentially increases.

In this paper, we propose a dimensionality reduction method with the TT subspace embedded, based on the Frobenius norm to measure the distance. We name our method tensor-train parameterization for ultra dimensionality reduction (TTPUDR). It enables the spatial relational information in the tensor to be efficiently and effectively processed and scrutinized, especially when the tensor is with a large number of modes or dimensions. Even for extreme outliers, the results in terms of accuracy and storage efficiency still appear to be satisfactory. In particular, the storage efficiency is higher than the existing dimensionality reduction methods such as PCA, LPP and TLPP. The main contribution of the paper lies in the following:
\begin{enumerate}
\item The proposed TTPUDR is the first example, intending to fill the research gap mentioned above. The embedded TT subspace can preserve spatial relations in multi/high-dimensional data and achieve lower storage complexity than the Tucker-based subspace in \cite{TLPP}. 

\item We propose to use the Frobenius norm (F-norm) in the tensor-train LPP (TTLPP) objective function to greatly reduce the sensitivity to the outliers, especially the extreme ones, and consider the spatial relations sufficiently. 

\item An efficient algorithm is proposed so that TTPUDR is sustainable and executable for ultra-dimensional data. This is a significant improvement over the approximated pseudo PCA implemented in the state-of-the-art dimensionality reduction method- tensor train neighborhood preserving embedding (TTNPE) \cite{WangAggarwalAeron2018}. 

\item A number of numerical experiments have been conducted on several real-world datasets. Its performance on these datasets is precisely consistent with the stated contributions and advantages. 
\end{enumerate}

\section{Related Work}\label{Sec:2}
As aforementioned, there are a great number of tensor decomposition methods investigating spatial relations in multi-dimensional data, i.e., tensors \cite{CPDecomp,TuckerDecomp,TTDecomp,TensorDecomp1,TensorDecomp2}. The tensor-train (TT) decomposition is relatively most 
efficient and effective among the above 3 classical methods. 

To preserve spatial information within tensors in the dimensionality reduction methods, \cite{TLPP} introduces the Tucker LPP (TLPP) which is LPP based on the Tucker decomposition to analyze the high-dimensional data and 
has the exponential increase in storage complexity as the number of modes increases. 

The other existing dimensionality reduction method which embeds the TT subspace, is the tensor train neighbourhood preserving embedding (TTNPE) \cite{WangAggarwalAeron2018}. 
TTNPE solves the exponential explosion on the complexity with the number of modes increasing. However, its robustness to the extreme outliers remains as a concern.
Therefore, a dimensionality reduction method for tensors with a large number of modes or dimensions is demanded to propose on the TT subspace and 
the capability of reducing the sensitivity to the extreme outliers. Our method TTPUDR is thus 
developed with all the aspects.  

\subsection{Preliminaries}
Before introducing the TT decomposition and LPP, the ground definitions, the notations and tensor operations are specified. In this paper, we do not distinguish the dimensions of a tensor and its modes. A classic vector is a tensor of mode 1 or 1-order tensor. Similarly, a matrix is a tensor of mode 2, i.e., a 2-order tensor; and a 3-order tensor can be viewed as a data cubic with three modes.

As the tradition, we denote the scalars by lower-case letters, such as $a$;  the vectors by the bold lower-case letters, for instance, $\mathbf{x}$; the matrices as the bold capital letters, for example, $\mathbf{S}$. They are all examples of tensors. In general, we use the calligraphic capital letters as the notations for tensors, e.g., $\mathscr{X} \in \mathbb{R}^{I_1\times I_2\times \cdots \times I_n}$ being an $n$-order tensor of dimension $I_i$ at mode $i$. 

Tensor contraction is defined as the multiplication of tensors along their compatible modes. 
Let $\mathscr{X}\in \mathbb{R}^{I_1\times I_2\times I_3\times \cdots \times I_n}$ and $\mathscr{Y}\in \mathbb{R}^{J_1\times J_2\times J_3\times \cdots \times J_m}$. The tensor contraction is defined as
\begin{align}
    \mathscr{Z}=\mathscr{X}\times_{\tilde{p}}^{\tilde{q}}\mathscr{Y}\label{Eq:contraction}
\end{align}
where $\tilde{p}\subseteq p=\{1,\cdots,n\}$ and $\tilde{q}\subseteq q=\{1,\cdots,m\}$ are subsets satisfying 
$\tilde{p}=\{k|I_k=J_k \}$ and $\tilde{q}=\{k|I_k=J_k \}$, 
respectively. 
The tensor contraction merges two tensors along the modes with the equal sizes, per se, and $\mathscr{Z}\in \mathbb{R}^{\times_{k\in \tilde{p}^c}I_k \times_{k\in \tilde{q}^c}J_k}$.

We denote the left unfolding operation \cite{WangAggarwalAeron2018} of $\mathscr{X}\in \mathbb{R}^{I_1\times I_2\times I_3\times \cdots\times I_n\times R_n}$ as the matrix $\mathbf{L}(\mathscr{X})\in \mathbb{R}^{I_1 I_2I_3\cdots I_n \times R_n}$ where the last mode of the tensor becomes the column indices of the left unfolding matrix and the rest of the modes are the row indices. Similarly, for the right unfolding operation, denoting it as 
$\mathbf{R}(\mathscr{X})\in \mathbb{R}^{I_1\times I_2\cdots I_nR_n}$. 
Also, the vectorization of a tensor is denoted by 
$\mathbf{V}(\mathscr{X})\in \mathbb{R}^{I_1 I_2\cdots I_n R_n}$. The F-norm of a tensor can be defined as the $\ell_2$-norm of its vectorization, i.e., $\|\mathscr{X}\|_F = \|\mathbf{V}(\mathscr{X})\|_2=\sqrt{\sum_{i_1=1}^{I_1}\sum_{i_2=1}^{I_2}\cdots\sum_{i_n=1}^{I_n}\sum_{r_n=1}^{R_n}x_{i_1,i_2,\cdots,i_n,r_n}^2}$, which considers all the elements $x_{i_1,i_2,\cdots,i_n},i_1=1,\cdots,I_1,\cdots,i_n=1,\cdots,I_n,r_n=1,\cdots,R_n$ as an entire group and preserves the general spatial relations between elements. Besides $\ell_1$-norm of a tensor is computed as $\|\mathscr{X}\|_1=\|\mathbf{V}(\mathscr{X})\|_1=\sum_{i_1=1}^{I_1}\sum_{i_2=1}^{I_2}\cdots\sum_{i_n=1}^{I_n}\sum_{r_n=1}^{R_n}|x_{i_1,i_2,\cdots,i_n,r_n}|$ which treats each elements separately and can probably cause the spatial information loss.

\subsection{Tensor-Train Decomposition}\label{Sec:2:1}
The tensor-train (TT) decomposition is designed for large-scale data analysis \cite{TTDecomp}. 
It can achieve a simpler implementation than the tree-type decomposition algorithms \cite{Tree-typeDecomp} which are developed to reduce the storage complexity and avoid the local minima.

The TT decomposition assumes a special structure of a tensor subspace where an $n$-order tensor is expressed as the contraction of a series of $n$ 3-order tensors. Specifically speaking, any element of an $n$-order tensor $\mathscr{Y}\in \mathbb{R}^{I_1\times I_2\times I_3\times \cdots \times I_n}$ is formed as follows, 
\begin{align}
    \mathscr{Y}(i_1, i_2, \cdots, i_n)=&\mathscr{U}_1(:,i_1,:)\mathscr{U}_2(:,i_2,:)\cdots\mathscr{U}_k(:,i_k,:)\notag\\
    &\cdots\mathscr{U}_{n-1}(:,i_{n-1},:) \mathscr{U}_n(:,i_n,:) \label{Eq:TTT}
\end{align}
where $\mathscr{U}_1\in \mathbb{R}^{1\times I_1\times R_1}$, $\mathscr{U}_k \in \mathbb{R}^{R_{k-1}\times I_k\times R_{k}}$ ($1<k<n$), and $\mathscr{U}_n\in\mathbb{R}^{R_{n-1}\times I_n\times1}$. $R_k$ ($k=1,2,\cdots,n-1$) are the tensor ranks. Let $R=\max\{R_1, R_2,\cdots, R_{n-1}\}$ and $I=\max\{I_1, I_2,\cdots, I_n\}$. Thus, the storage complexity is $\mathcal{O}(nIR^2)$ for the TT decomposition.

For most of the applications, in order to achieve the computational efficiency and be less information redundant, the researchers often restrict the tensor ranks to be smaller than the size of their corresponding tensor mode, i.e., $R_k <I_k$ for $k=1,2,\cdots,n-1$ \cite{WangAggarwalAeron2018}.

\subsection{Locality Preserving Projections}\label{Sec:2:2}
Locality preserving projections (LPP) \cite{LPP} is to explore and preserve local information of data in the projected lower dimensional space,  while 
the conventional principal component analysis (PCA) \cite{PCA} favours maintaining global information in data. 

Given a set of vectorial training data $\{\mathbf{x}_i\}^N_{i=1} \subset \mathbb{R}^P$ and an affinity matrix of locality similarity $\mathbf S = [s_{ij}]$,  
LPP intends to seek for a linear projection $\mathbf A$ from $\mathbb{R}^P$ to $\mathbb{R}^p$ such that the following optimization problem is solved to minimize the locality preserving criterion set as the objective function.
\begin{align}
&\min_{\mathbf A}\sum_{i,j}\|\mathbf A^\top \mathbf x_i -\mathbf A^\top \mathbf x_j\|_2^2 s_{ij}\notag\\
&\text{s.t.} \quad \mathbf{A}^\top\mathbf{X}\mathbf{D}\mathbf{X}^\top\mathbf{A}=\mathbf{I}
\label{LPP:obj}
\end{align}
The widely used affinity $\mathbf S = [s_{ij}]$  is based on 
the graph of the neighborhood information in the data as follows \cite{LPP}.
\begin{align}
    s_{ij}=
    \begin{cases}
    e^{-\frac{||\mathbf{x}_i-\mathbf{x}_j||^2_F}{t}},& \text{if } \mathbf{x}_i\in \mathcal{N}_k(\mathbf{x}_j)\quad\text{or } \mathbf{x}_j \in \mathcal{N}_k(\mathbf{x}_i)\notag\\
    0,              & \text{otherwise}\notag
\end{cases} 
\end{align}
where $t\in \mathbb{R}_+$ is a positive parameter and $\mathcal{N}_k(\mathbf{x})$ denotes the $k$-nearest neighborhood of $\mathbf x$.

Denote $\mathbf{X}=[\mathbf{x}_1, \mathbf{x}_2,\cdots,\mathbf{x}_{N-1},\mathbf{x}_N]$. 
The LPP problem \eqref{LPP:obj} indeed can be converted to the following generalized eigenvalue problem to solve the eigenvalues $\lambda$ and eigenvectors $\mathbf{a}$.

\begin{align}
    \mathbf{X}\mathbf{L}\mathbf{X}^\top \mathbf{a}=\lambda\mathbf{X}\mathbf{D}\mathbf{X}^\top\mathbf{a} \label{Eq:Eign}
\end{align}
where $\mathbf L = \mathbf D - \mathbf S$ and $\mathbf D$ is a diagonal matrix consisting of the row sum of $\mathbf S$. The columns of the final mapping $\mathbf A$ consist of the generalized eigenvectors $\mathbf{a}$ in Equation~\eqref{Eq:Eign}, corresponding to the smallest $p$ eigenvalues $\lambda$'s. 

LPP is a classical dimensionality reduction method and has been applied in many real cases, for example, computer vision \cite{LPPCV} 
. It captures the local information among the data points and reduces more sensitivity to the outliers than PCA. However, we do observe the following shortcomings of LPP:
\begin{enumerate}
\item LPP is designed for vectorial data. When it is applied to 
multi-dimensional data, i.e, tensors, there exists potential loss of spatial information. 
The existing tensor locality preserving projections, i.e., the Tucker LPP (TLPP) \cite{TLPP} 
embeds the tensor space with a high storage complexity at $\mathcal{O}(nIR+R^n)$. 
\item Theoretically, LPP cannot work for the cases where the data dimension is greater than the number of samples. Although this can be avoided by a trick in which one first projects the data onto its PCA subspace, then implements LPP in this subspace\footnote{ \url{http://www.cad.zju.edu.cn/home/dengcai/Data/code/LPP.m}}, this would not work well for ultra-dimensional data with a fairly large dataset as a singular value decomposition (SVD) becomes a bottleneck.    
\end{enumerate}

The TT decomposition with a smaller storage complexity at $\mathcal{O}(nIR^2)$ 
has been recently applied in the tensor train neighborhood preserving embedding (TTNPE) 
\cite{WangAggarwalAeron2018,WangAggarwalAeron2017}. Nevertheless, the actual algorithm in TTNPE is only implemented as a TT approximation to the pseudo PCA. 
To the best of our knowledge, there is no existing dimensionality reduction method which can directly process the tensor data with less storage complexity, i.e., using the TT decomposition in algorithms.

\section{Methodology}\label{Sec:4}
In this section, we propose the tensor-train parameterization for ultra dimensionality reduction (TTPUDR) to fill the research gap aforementioned in Section~\ref{Sec:1}. 
The learning procedure is presented in detail with a summary in the form of pseudo code.

Consider a tensor-train (TT) $\widetilde{\mathscr{U}} =\mathscr{U}_1 \times^{1}_{3}\mathscr{U}_2\times^{1}_{3}\cdots\times^{1}_{3}\mathscr{U}_n$ where $\mathscr{U}_k\in \mathbb{R}^{R_{k-1}\times I_k\times R_k}$, $R_0=1$ and $R_n=I_1 I_2\cdots I_n$. For a given set of tensor data $\{\mathscr{X}_i\}^N_{i=1}\subset \mathbb{R}^{I_1\times I_2\cdots \times I_n}$, we project $\mathscr{X}_i\in \mathbb{R}^{I_1\times I_2\times\cdots\times I_n}$ to the vector $\mathbf{t}_i\in \mathbb{R}^{R_n}$ by a TT parameterized mapping defined as, 
\[
\mathbf{t}_i=\mathbf{L}^\top(\widetilde{\mathscr{U}}) \mathbf{V}(\mathscr{X}_i)
\]
where $R_n$ now is the number of components or the dimension of $\mathscr{X}_i$.
Denote by $\mathbf S = [s_{ij}]$ the similarity based on the graph of the neighborhood of tensor data, which may be defined as used in LPP \cite{LPP} introduces in Section~\ref{Sec:2}. 
To increase the model robustness towards extreme data outliers and preserve the spatial relations, 
we design the TTPUDR by modifying the LPP formulation as the following optimization problem using the Frobenius norm objective function instead of applying the squared Frobenius norm or the $\ell_1$-norm, 
\begin{align} 
\min_{\mathscr{U}_1, \mathscr{U}_2,\cdots, \mathscr{U}_n}\quad &\frac{1}{2}\sum_{i,j=1}^N\|\mathbf{L}^\top(\widetilde{\mathscr{U}})\mathbf{V}(\mathscr{X}_i) -\mathbf{L}^\top(\widetilde{\mathscr{U}})\mathbf{V}(\mathscr{X}_j)\|_F s_{ij}  \notag\\ 
 \text{s.t.}\quad\mathbf{L}^\top(\mathscr{U}_k)&\mathbf{L}(\mathscr{U}_k)=\mathbf{I}_{R_k} \quad   \;\;\; \forall\; k=1,\cdots n.
\label{TTPUDROptProb}
\end{align}

The TT decomposition based parameterization for the mapping tensor  
can preserve or learn the spatial relation in tensor data $\mathscr{X}_i$. 
However, using the F-norm in Problem~\eqref{TTPUDROptProb} makes it more difficult to solve the problem of TTPUDR.  

We propose to use a splitting and iterative way to solve the problem. 
For this purpose, we define
\begin{align}
    \widetilde{s}_{ij}=\frac{s_{ij}}{\|\mathbf{L}^\top(\widetilde{\mathscr{U}})\mathbf{V}(\mathscr{X}_i)-\mathbf{L}^\top(\widetilde{\mathscr{U}})\mathbf{V}(\mathscr{X}_j)\|_F}\label{tildeS}
\end{align}
which is a function of the tensor cores $\mathscr{U}_1,\mathscr{U}_2,\cdots,\mathscr{U}_n$. Then we rewrite 
Problem~\eqref{TTPUDROptProb} 
in terms of  the squared F-norm as follows 
\begin{align} 
&\min_{\mathscr{U}_1, \mathscr{U}_2,\cdots, \mathscr{U}_n}\quad \frac{1}{2}\sum_{i,j=1}^N\|\mathbf{L}^\top(\widetilde{\mathscr{U}})\mathbf{V}(\mathscr{X}_i)-\mathbf{L}^\top(\widetilde{\mathscr{U}})\mathbf{V}(\mathscr{X}_j)\|^2_F \widetilde{s}_{ij}  \notag\\ 
&\text{s.t.}\quad\mathbf{L}^\top(\mathscr{U}_k)\mathbf{L}(\mathscr{U}_k)=\mathbf{I}_{R_k} \quad   \;\;\; \forall\; k=1,\cdots n.
\label{TTPUDRReal}
\end{align}

Problem~\eqref{TTPUDRReal} seems to be an LPP problem. However, it is not because the modified affinity $\widetilde{s}_{ij}$ is a function of parameters $\{\mathscr{U}_1, \mathscr{U}_2,\cdots, \mathscr{U}_n\}$. 
We solve it in the following way. Suppose Problem~\eqref{TTPUDRReal} is being solved by an iterative optimization algorithm. We use the current parameter values to calculate $\widetilde{s}_{ij}$ according to Equation~\eqref{tildeS} and then fix all $\widetilde{s}_{ij}$ to solve Problem~\eqref{TTPUDRReal}. This alternative procedure can continue until convergence.

To efficiently solve Problem~\eqref{TTPUDRReal} while $\widetilde{s}_{ij}$ fixed, 
we follow an alternating procedure for solving each tensor core $\mathscr{U}_k$ while the rest are fixed. 
Overall, we solve the TT parameters, i.e., tensor cores, and update the neighborhood graph $\tilde{\mathbf{S}}$ alternately. 
This learning procedure terminates when the solution converges. 

In optimizing each tensor core $\mathscr{U}_k$, 
we find that the strategy in \cite{WangAggarwalAeron2017}  
involves manipulating a matrix $\mathbf Z  \in \mathbb{R}^{I_1 I_2 \cdots I_n\times I_1 I_2 \cdots I_n}$, which is forbidden when data are ultra-dimension or high-order tensors. By taking the commutative property of the tensor contraction operation, we propose a new strategy which largely speeds up the calculation.

To describe the new algorithm, we define 
\begin{align}
    \mathscr{T}_1(k)=\mathscr{U}_1\times_3^1\cdots\times_3^1\mathscr{U}_{k-1}\in \mathbb{R}^{I_1\times I_2\times\cdots I_{k-1}\times R_{k-1}} \label{T1},\\
    \mathscr{T}_n(k)=\mathscr{U}_{k+1}\times_3^1\cdots\times_3^1\mathscr{U}_{n}\in \mathbb{R}^{R_k\times I_{k+1}\times\cdots \times I_{n}\times R_{n}}\label{Tn}
\end{align}
where $1\leq k\leq n$ but $\mathscr{T}_1(1)$ and $\mathscr{T}_n(n)$ are not defined. Let $\mathscr{X}$ be the $(n+1)$-order data tensor whose mode-$(n+1)$ stacks along the data samples, i.e., $\mathscr{X}\in\mathbb{R}^{I_1\times I_2\times \cdots \times I_n\times N}$. Then define the partially transformed tensor, for $1<k<n$ of size $R_{k-1}\times R_k\times R_n\times I_k\times N$, 
\[
\mathscr{Y}_k=(\mathscr{X}\times^{1,2,...,k}_{1,2,..., k}\mathscr{T}_1(k)) \times^{2, 3, ..., n+1-k}_{2, 3, ..., n+1-k} \mathscr{T}_n(k), 
\]
and, for $k=1$,
\[
\mathscr{Y}_1= \mathscr{X}\times^{2,...,n}_{2,..., n}\mathscr{T}_n(1) \in \mathbb{R}^{R_1\times R_n\times I_1\times N},
\]
and, for $k=n$,
\[
\mathscr{Y}_n= \mathscr{X}\times^{1,...,n-1}_{1,..., n-1}\mathscr{T}_1(n) \in \mathbb{R}^{R_{n-1}\times I_n\times N}.
\]

Finally, the optimization problem~\eqref{TTPUDRReal} for TTPUDR is transformed to the following subproblems, respectively: 

\textbf{Solving for $\mathscr{U}_1$:}  For each $1\leq r_n\leq R_n$, take the slice $\mathscr{Y}_1(:,r_n, :, :)$ and reshape it as a matrix $\mathbf Y_1(r_n)$ of size $(R_1I_1)\times N$, and form the matrix $\mathbf H_1 = \sum^{R_n}_{r_n=1} \mathbf Y_1(r_n) \widetilde{\mathbf L} \mathbf Y_1(r_n)^\top$ of size $(R_1I_1)\times (R_1I_1)$. Then $\mathscr{U}_1$ is solved by 
\begin{align}
\min_{\mathscr{U}_1}\mathbf V(\mathscr{U}_1)^\top \mathbf H_1 \mathbf V(\mathscr{U}_1),\; \text{s.t.} \;\mathbf{L}^\top(\mathscr{U}_1)\mathbf{L}(\mathscr{U}_1)=\mathbf{I}_{R_1}   
    \label{TTPUDR1}
\end{align}

\textbf{Solving for $\mathscr{U}_k$ ($1<k<n$):}  For each $1\leq r_n\leq R_n$, take the slice $\mathscr{Y}_k(:,:, r_n, :, :)$ and reshape it as a matrix $\mathbf Y_k(r_n)$ of size $(R_{k-1}I_kR_k)\times N$, and form the   matrix $\mathbf H_k = \sum^{R_n}_{r_n=1} \mathbf Y_k(r_n) \widetilde{\mathbf L} \mathbf Y_k(r_n)^\top$ of size $(R_{k-1}I_kR_k)\times (R_{k-1}I_kR_k)$. Then $\mathscr{U}_k$ is solved by
\begin{align}
\min_{\mathscr{U}_k}\mathbf V(\mathscr{U}_k)^\top \mathbf H_k \mathbf V(\mathscr{U}_k), \;\text{s.t.} \;  \mathbf{L}^\top(\mathscr{U}_k)\mathbf{L}(\mathscr{U}_k)=\mathbf{I}_{R_k}.  
    \label{TTPUDR2}
\end{align}

\textbf{Solving for $\mathscr{U}_n$:} Reshape $\mathscr{Y}_n$ to the matrix $\mathbf Y_n$ of size $(R_{n-1}I_n)\times N$, and form the matrix
$\mathbf H_n =  \mathbf Y_n  \widetilde{\mathbf L} \mathbf Y^\top_n$. Then solve $\mathscr{U}_n$ satisfying $\mathbf{L}^\top(\mathscr{U}_n)\mathbf{L}(\mathscr{U}_n)=\mathbf{I}_{R_n}$ by 
\begin{align}
\min_{\mathscr{U}_n}\text{trace}(\mathbf L^\top(\mathscr{U}_n) \mathbf H_n \mathbf L(\mathscr{U}_n)).  
    \label{TTPUDR3}
\end{align}

Each problem in \eqref{TTPUDR1} -- \eqref{TTPUDR3} is an optimization problem over Stiefel manifolds of small dimensions. They can be efficiently solved by manifold optimization package such as ManOpt (\url{http://www.manopt.org}).
\begin{algorithm}[t]
\caption{Optimization Algorithm for TTPUDR}\label{alg:euclid}
\hspace*{\algorithmicindent} \textbf{Input}: $\mathscr{X}=\{\mathscr{X}_i\}^N_{i=1} \subset \mathbb{R}^{I_1\times I_2\times \cdots\times I_n}$,
the original neighbourhood graph $\mathbf{S}$, 
and the number of maximum iterations $Iter$. \\
 \hspace*{\algorithmicindent} \textbf{Output}: Optimal tensor cores $\mathscr{U}_1, \mathscr{U}_2,\cdots,\mathscr{U}_n$ for the tensor train.
\begin{algorithmic}[1]
\State Initialize the tensor cores $\mathscr{U}_k \in\mathbb{R}^{R_{k-1}\times I_k\times R_{k}}$ for $k=1,\cdots,n$. For $\mathscr{U}_1\in \mathbb{R}^{R_{0}\times I_1\times R_{1}}$, $R_0=1$ and for $\mathscr{U}_n\in \mathbb{R}^{R_{n-1}\times I_n\times R_{n}}$,  {$R_n=1,2,\cdots,I_1I_2\cdots I_n$.}
\For{$m=1:Iter$}
\State Calculate $\widetilde{\mathbf S} = [\widetilde{s}_{ij}]$ according to Equation~\eqref{tildeS} and prepare $\widetilde{\mathbf L} = \widetilde{\mathbf D} - \widetilde{\mathbf S}$;
\State 
\For{$k=1:n$}
\If{$k=1$}
\State Form the problem~\eqref{TTPUDR1} by calculating $\mathbf H_1$ and obtain $\mathscr{U}_1$ by solving the problem;
\ElsIf{$k=2,\cdots, n-1$}
\State Form the problem~\eqref{TTPUDR2} by calculating $\mathbf H_k$ and obtain $\mathscr{U}_k$ by solving the problem;
\Else
\State Form the problem~\eqref{TTPUDR3} by calculating $\mathbf H_n$ and obtain $\mathscr{U}_n$ by solving the problem;
\EndIf
\EndFor
    \If{converge} 
    \State $break$
    \EndIf
\EndFor\label{euclidendwhile}
\State \textbf{return} $\mathscr{U}_1,\cdots, \mathscr{U}_n$
\end{algorithmic}
\end{algorithm}
To sum up, the pseudo code for the entire learning process of TTPUDR is presented in Algorithm~\ref{alg:euclid}. Note that there has not been any perfect theoretical proof of the convergence of TTPUDR, but it still achieves the convergence empirically as shown in the experiments in Section~\ref{Sec:5}.

\textit{Remark 1:} We have added the orthogonal constraints $\mathbf{L}^\top(\mathscr{U}_k)\mathbf{L}(\mathscr{U}_k)=\mathbf{I}_{R_k}$ in Problems~\eqref{TTPUDR1} - \eqref{TTPUDR3}. These constrained conditions make sure that the dimensionality reduction mapping $\mathbf{E}=\mathbf{L}(\mathscr{U}_1\times_3^1\mathscr{U}_2\times_3^1\cdots\times_3^1\mathscr{U}_n)$ consists of orthogonal columns, by referring to Lemma 2 in \cite{WangAggarwalAeron2018}. To ease the optimization on the Stiefel manifold in Problems~\eqref{TTPUDR1} and \eqref{TTPUDR2}, we can replace the orthogonal condition by $\mathbf{V}^\top(\mathscr{U}_k)\mathbf{V}(\mathscr{U}_k)=1$ ($1\leq k<n$), resulting in an eigenvalue problem. However, the overall orthogonality will be lost.

\textit{Remark 2:} Problem \eqref{TTPUDR3} is quite different from Problems \eqref{TTPUDR1} and \eqref{TTPUDR2}. Problem \eqref{TTPUDR3} is equivalent to the eigenvalue problem of $\mathbf H_n$. 

\textit{Remark 3:} The algorithm can be used for dimensionality reduction for ultra-dimensional vectorial data. For example, suppose that the dimension of vector data is $D = I_1\times I_2\times \cdots \times I_n$, then we can seek for the dimensionality reduction mapping in terms of TT parameterization. This makes dimensionality reduction possible for ultra-dimensional data.

\section{Experiments}\label{Sec:5}

To validate the proposed TTPUDR method,
the experiments on facial recognition and remote sensing are demonstrated in this section. 
The results are compared with the classical methods and its related methods, i.e., PCA \cite{PCA} and LPP \cite{LPP}.  All the experiments are conducted on the Windows 10 system with the memory at 128GB and the Intel Core i7 6950X processor for 25M cache and up to 3.50 GHz, with Matlab 2018a version.  
\subsection{Data Description}
The performance of the TTPUDR method is studied through numerical experiments on two high-dimensional datasets from two publicly available databases: the Extended Yale B \cite{YaleB} and the Northwest Indiana’s Indian Pines by the Airborne Visible/Infrared Imaging Spectrometer (AVIRIS) sensor in 1992 \cite{IndianaData}. The first two experiments are conducted on the original datasets from these two databases, whereas the third experiment aims to investigate the robust property of TTPUDR on extreme outliers. Therefore, we add the $10\%$ block noises to the Extended Yale B dataset.

The Extended Yale B dataset is on facial of 38 individuals. Each individual has 9 positions and 64 near frontal-face images, resulting in a total of $21888$ images. Each image has been resized to $32\times 32$ pixels. After conducting the rearrangements and removing the missing values, 
the final number of images is $2414$.

In terms of the Northwest Indiana's Indian Pine (Indiana) dataset, it is collected based on the Indian Pines test site in North-western Indiana and contains $145\times145$ pixels and $224$ spectral reflectance bands in the wavelength range $0.4–2.5 \times 10^{(-6)}$ meters. Similar to what is in the Extended Yale B dataset, we choose $200$ spectral reflectance bands and $10366$ pixel locations by eliminating the missing values and the water absorption.

For the noised Extended Yale B dataset, we add the block noise to $10\%$ and $20\%$ of the images for each. The noises are generated as either the minimum value or the maximum value of the Extended Yale B dataset as either $0$ or $255$, whereas the general pixel values are from $9$ to $115$. They are added as $4\times 4$ blocks to images, which are salt and pepper noises. Their locations are both predefined and random. 
This dataset is designed to examine the robustness
of TTPUDR to the extreme outliers.  

To investigate the capability of capturing the spatial structure information, we select the first two datasets in the three datasets above. In these two datasets (no noises added), $60\%$ of the data are considered as the training set and $40\%$ of the data are regarded as the test set.  
Then to test the robustness and further scrutinize the ability of TTPUDR to process the ultra high-dimensional data, we only utilize the third noised dataset, where $60\%$ and $20\%$ of the data are treated to be the training set and $40\%$ and $80\%$ of the data are set as the test set, respectively. In the case of the noised dataset, the extreme outlier noises are added at $10\%$ and $20\%$ among the training data accordingly. 

\subsection{Benchmark and Comparison Criteria} 
The experiments are designed to evaluate the capability to analyze the structured high-dimensional data and the robustness to the extreme outliers of TTPUDR. 
We compare its performance with existing methods such as PCA and LPP for compatible cases. Note that we are unable to compare with TTNPE since its publicly available program itself is not executable due to its extreme computational complexity. For TLPP, the same issue also exists. For both PCA and LPP, we use the implementation in \url{https://lvdmaaten.github.io/drtoolbox/}.

For the classification performance, we use the data after dimensionality reduction as the new features for each object and conduct a classifier fitting.  The 1-nearest neighborhood (1NN) classifier is used in our experiments. 
The evaluation criteria are the overall accuracy (OA), the average accuracy (AA), 
and Kappa coefficient (KC) for the number of reduced dimensions from 2 to 30, i.e., $R_n=2,\cdots,30$. Specifically, these criteria are computed as 
\begin{align}
    OA &= \frac1{T}\sum_{c=1}^C TP_{c }, \quad\quad\quad AA = \frac1{C}\sum_{c=1}^C \frac{TP_{c}}{TP_{c}+FP_{c}}, \nonumber\\
    KC&= \frac{OA-\frac1{C^2}\sum_{c=1}^C(TP_{c}+FP_{c})\times(TP_{c}+FN_{c})}
    {1-\frac1{C^2} \sum_{c=1}^C(TP_{c}+FP_{c})\times(TP_{c}+FN_{c})}. \notag 
\end{align}
where $C$ is the total number of classes and $T$ is the number of the test data points. 

For robustness to outliers, the evaluation criteria are on the accuracy itself and the convergence speed of the accuracy, for the different proportion of outliers 
at $10\%$ and $20\%$. 
Furthermore, the convergence analysis is conducted based on the four cases mentioned above, but only the case with the fastest convergence speed for TTPUDR is disclosed and compared with the same three methods across all the iterations on the corresponding feature number for TTPUDR. 

\subsection{Results and Findings}\label{Sec:5:3}

As aforementioned, the experiments on the Indiana dataset and the Extended Yale B dataset are to examine how TTPUDR can capture the spatial information in the high-dimensional data. We also apply the noised Extended Yale B dataset to examine the robustness of TTPUDR. In the first set of experiments, the dimension of the training data is smaller than the number of samples. Another set of experiments on the noised Extended Yale B is intended to further evaluate this ability of TTPUDR on ultra high-dimensional data and its robustness to extreme outliers.

\subsubsection{Parameter Compression Capability}
In the case with spatial information capturing, the dimension of the data is smaller than the number of samples for the training set. In other words, the assumption of LPP is not violated on the dimension size and the number of samples. On each method for each dataset, we have executed them for 150 iterations, i.e., 10 shuffles of random samples with 15 iterations for each sample. Firstly, the results for the Indiana dataset is presented in Table~\ref{Table1}. 

\begin{table}[h!]
\centering
\begin{tabular}{ |p{1cm}||p{1.5cm}|p{1.5cm}|p{1.5cm}|  }
 \hline
 \multicolumn{4}{|c|}{Results from the Indiana Dataset} \\
 \hline
 & PCA&LPP&TTPUDR\\
 \hline
 OA&    $0.7907$ &  $0.7810$&    $0.7101$\\
 AA&    $0.7983$&  $0.8191$&    $0.7427$\\
 KC&    $0.7613$&  $0.7497$&  $0.6690$\\
 \hline
\end{tabular}
\caption{Comparison of evaluation criteria under TTPUDR, LPP and PCA on the Indiana dataset.}
\label{Table1}
\end{table}

For the fair comparison, the number of neighbors and the parameter $t$ are set as $4$ and $0.02$ respectively to construct the affinity matrix of locality similarity $\mathbf{S}$ for both LPP and TTPUDR. The sizes of tensor cores in TTPUDR are $1\times4\times3$, $3\times5\times4$ and $4\times10\times R_n$ with $R_n$ from 2 to 30 as the number of features. The total numbers of model parameters are from 152 to 1272, verse 200 to 6000 for PCA and LPP. Here we only present the case with $R_n=24$ which is randomly selected from 2 to 30. 
The values in each cell of the table are the means of the 10 randomnesses. As this dataset has a larger number of samples and a smaller number of dimensions, the performance of the proposed TTPUDR is less competitive to that of PCA and LPP. On average, OA, AA and KC values under TTPUDR are 10\% smaller than those under PCA and LPP.

A similar experiment for the Extended Yale B dataset can be demonstrated in Table~\ref{Table2}.  
\begin{table}[h!]
\centering
\begin{tabular}{ |p{1cm}||p{1.5cm}|p{1.5cm}|p{1.5cm}|  }
 \hline
 \multicolumn{4}{|c|}{Results from the Extended Yale B Dataset} \\
 \hline
 & PCA&LPP&TTPUDR\\
 \hline
 OA&    $0.4461$ &  $0.4378$&    $0.7557$\\
 AA&  $0.4937$  &   $0.4491$&    $0.7731$\\
 KC&    $0.4312$ &  $0.4460$&    $0.7491$\\
 \hline
\end{tabular}

\caption{Comparison of evaluation criteria under TTPUDR, LPP and PCA on the Extended Yale B dataset.}
\label{Table2}
\end{table}

To compare TTPUDR with LPP fairly, the number of neighbors and the Heat kernel width parameter $t$ are set as $4$ and $0.5$ respectively to construct the affinity matrix of locality similarity $\mathbf{S}$ for both LPP and TTPUDR. The sizes of tensor cores in TTPUDR are $1\times4\times4$, $4\times8\times7$, $7\times4\times4$ and $4\times8\times R_n$ with $R_n$ from 2 to 30 as the number of features. The total numbers of model parameters are from 416 to 1312, verse 2048 to 30720 for PCA and LPP. In this case, we randomly choose $R_n=28$ to demonstrate. The numbers in Table~\ref{Table2} are also the best result of each method for each criterion. In this case, the results are based on $R_n=28$ features, i.e., dimensions. This case shows that TTPUDR performs better than both PCA and LPP. On average, these values are at least 66\% bigger under TTPUDR than LPP and PCA. The presented OA, AA and KC in the table are also the means of those across iterations. This result is not surprising as this dataset has a smaller sample size and a larger dimension than the Indiana dataset, which align with the characteristics of ultra-dimensionality under TTPUDR.

This set of experiments has demonstrated that the TTPUDR uses much fewer model parameters to achieve comparable performance for the classification tasks.

\subsubsection{Robustness}
\begin{figure}[!h]
\centering
\subfloat[]{%
\label{fig:ex3-a}%
\includegraphics[width=0.233\textwidth]{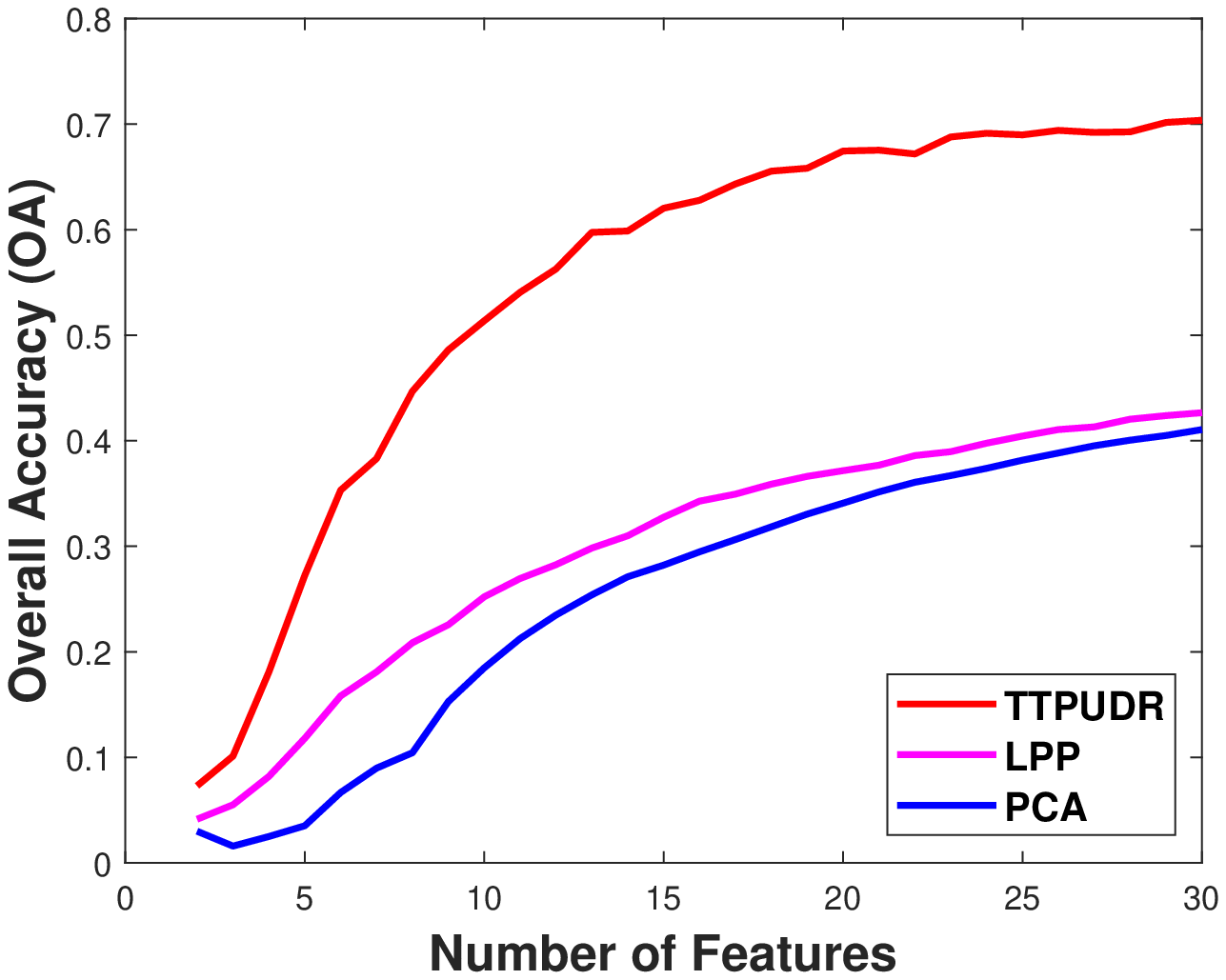}}%
\hspace{8pt}%
\subfloat[]{%
\label{fig:ex3-b}%
\includegraphics[width=0.233\textwidth]{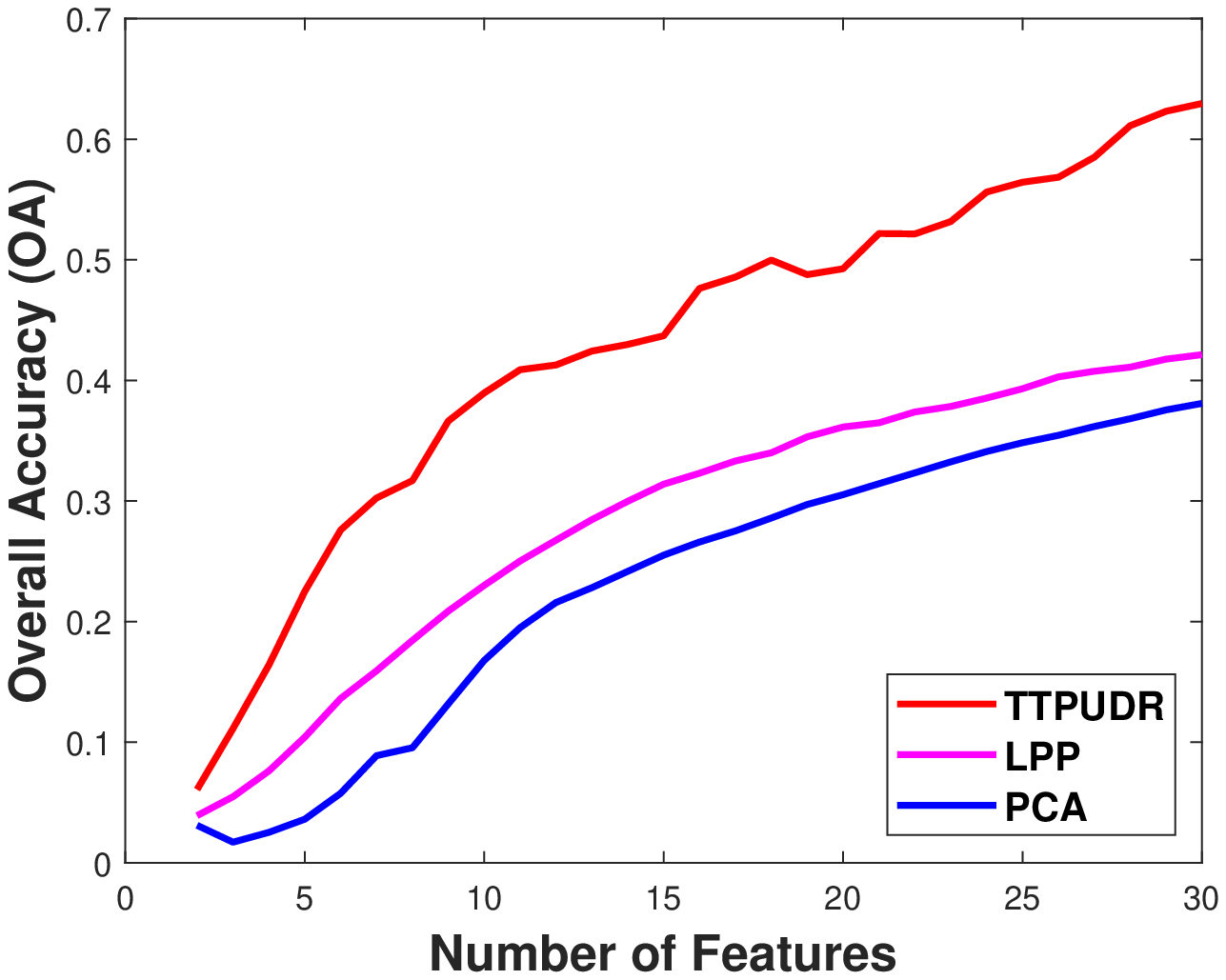}}  \\
\subfloat[]{%
\label{fig:ex3-c}%
\includegraphics[width=0.233\textwidth]{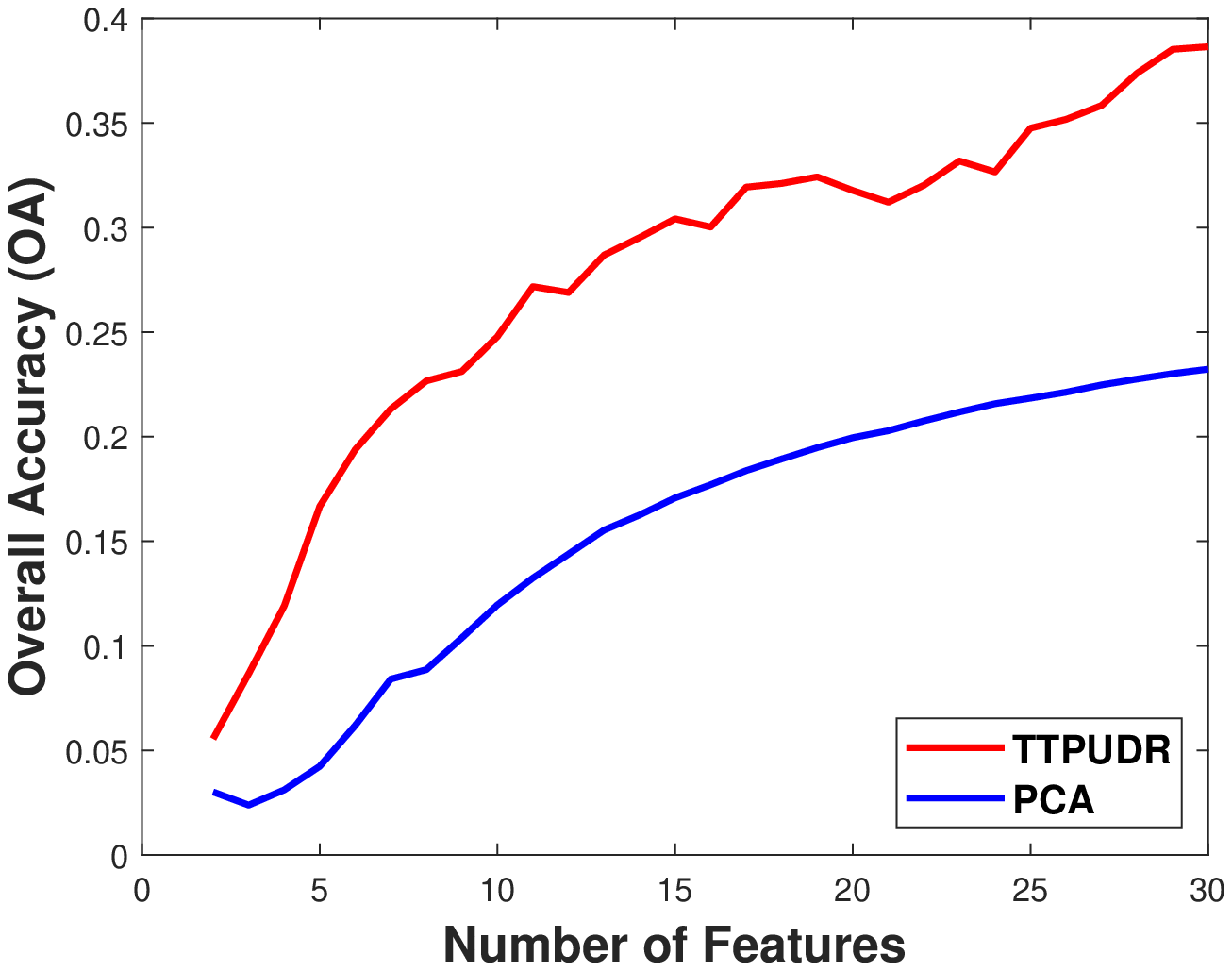}}%
\hspace{8pt}%
\subfloat[]{%
\label{fig:ex3-d}%
\includegraphics[width=0.233\textwidth]{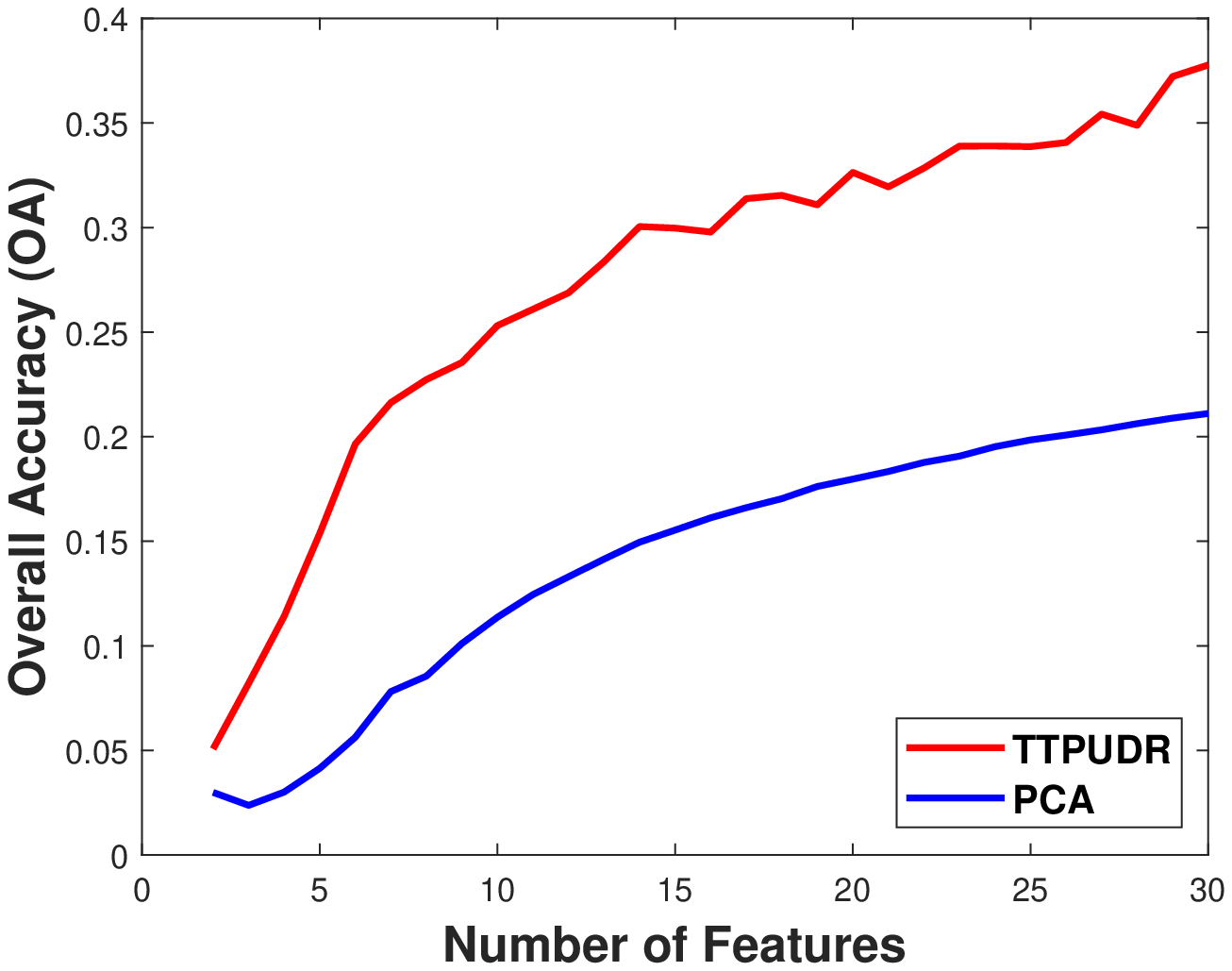}}%
\caption[A set of four subfigures.]{Comparison of overall accuracy (OA) for TTPUDR, LPP and PCA in the noised Extended Yale B dataset under \subref{fig:ex3-a} $60\%$ training data and $10\%$ of the noise;
 \subref{fig:ex3-b} $60\%$ training data and $20\%$ of the noise;
 \subref{fig:ex3-c} $20\%$ training data and $10\%$ of the noise; and,
 \subref{fig:ex3-d} $20\%$ training data and $20\%$ of the noise.}
\label{fig:ex3}%
\end{figure}

Following the parameter compression capability, we examine the robustness of TTPUDR with the noised Extended Yale B dataset.
The results are reported in Figure~\ref{fig:ex3}. For simplicity, we present OA for TTPUDR, LPP and PCA across dimensions, i.e., features from 2 from 30, since all the three methods have the best performance on this evaluation criterion than the other criteria. 

Figures~\ref{fig:ex3-a} and~\ref{fig:ex3-b} demonstrate the performance of TTPUDR, LPP and PCA with 60\% of training data with 10\% and 20\% of extreme outlier noise, respectively. From these Figures, it is evident that TTPUDR significantly outperforms LPP and PCA on the overall accuracy. In the case with $10\%$ of the noise, TTPUDR generally achieves better performance at a lower reduced dimensionality although this pace has slightly slowed down in the case of  
the $20\%$ of extreme outlier noises. 
Therefore, we can conclude that TTPUDR is capable of capturing sufficient information in the ultra high-dimensional data effectively and efficiently under a lower dimensionality. In both cases of noises, TTPUDR has better performance than both LPP and PCA. This shows that TTPUDR has significantly higher robustness to the extreme outliers due to its adopting the F-norm LPP objective. 

In Figures~\ref{fig:ex3-c} and~\ref{fig:ex3-d}, we show the results for the case of using $20\%$ training data, resulting 482 samples of 1024 dimensions. Since the number of dimensions is larger than the number of samples, the assumption of LPP is violated. Thus, LPP is not able to execute and there is no result available for LPP. However, TTPUDR can still operate and produce a more satisfactory OA compared with the other benchmark method, PCA. To sum up, TTPUDR has an excellent capability of processing and analyzing the spatial structural information in the ultra high-dimensional data effectively even with a really small number of training data. In terms of the robustness, TTPUDR also has a more preferable performance than the other executable method.

\section{Conclusions}\label{Sec:6}
This paper proposes a tensor-train parameterization for the ultra-dimensionality reduction algorithm. The dimensionality reduction mapping is tensorized to learn and preserve spatial information amongst multi-dimensional data and to increase model robustness towards extreme data outliers. This method has been successfully illustrated in two real datasets. The performance of the method is comparable with the existing methods with less parameters. It also outperforms other competitive models in the case of high-dimension-small-samples and large proportion of data with extreme noises. In the future research, we intend to expand it into a structure which can also capture and analyze the sequential relations in the time series tensor data.

\bibliographystyle{IEEEtran}

\begin{thebibliography}{10}
\providecommand{\url}[1]{#1}
\csname url@samestyle\endcsname
\providecommand{\newblock}{\relax}
\providecommand{\bibinfo}[2]{#2}
\providecommand{\BIBentrySTDinterwordspacing}{\spaceskip=0pt\relax}
\providecommand{\BIBentryALTinterwordstretchfactor}{4}
\providecommand{\BIBentryALTinterwordspacing}{\spaceskip=\fontdimen2\font plus
\BIBentryALTinterwordstretchfactor\fontdimen3\font minus
  \fontdimen4\font\relax}
\providecommand{\BIBforeignlanguage}[2]{{%
\expandafter\ifx\csname l@#1\endcsname\relax
\typeout{** WARNING: IEEEtran.bst: No hyphenation pattern has been}%
\typeout{** loaded for the language `#1'. Using the pattern for}%
\typeout{** the default language instead.}%
\else
\language=\csname l@#1\endcsname
\fi
#2}}
\providecommand{\BIBdecl}{\relax}
\BIBdecl

\bibitem{TensorFace}
M.~Vasilescu and D.~Terzopoulos, ``Multilinear analysis of image ensembles:
  Tensorfaces,'' in \emph{ECCV}, A.~Heyden, G.~Sparr, M.~Nielsen, and
  P.~Johansen, Eds., 2002, pp. 447--460.

\bibitem{TensorRecomd}
P.~Symeonidis, ``Matrix and tensor decomposition in recommender systems,'' in
  \emph{ACM RecSys}, 2016, pp. 429--430.

\bibitem{TensorSignal}
A.~Cichocki, D.~Mandic, A.~Phan, C.~Caiafa, G.~Zhou, Q.~Zhao, and L.~Lathauwer,
  ``Tensor decompositions for signal processing applications from two-way to
  multiway component analysis,'' \emph{arXiv:1403.4462}, 2014.

\bibitem{TensorFMRI}
C.~F. Beckmann and S.~M. Smith, ``Tensorial extensions of independent component
  analysis for multisubject {FMRI} analysis,'' \emph{Neuroimage}, vol.~25,
  no.~1, pp. 294--311, 2005.

\bibitem{WangAggarwalAeron2016}
W.~Wang, V.~Aggarwal, and S.~Aeron, ``Tensor completion by alternating
  minimization under the tensor train ({TT}) model,'' \emph{arXiv;1609.05587},
  2016.

\bibitem{TLPP}
G.~Dai and D.~Yeung, ``Tensor embedding methods,'' in \emph{AAAI}, 2006, pp.
  330--335.

\bibitem{WangAggarwalAeron2018}
W.~Wang, V.~Aggarwal, and S.~Aeron, ``Principal component analysis with tensor
  train subspace,'' \emph{arXiv:1803.05026}, 2018.

\bibitem{CPDecomp}
F.~L. Hitchcock, ``Multiple invariants and generalized rank of a p-way matrix
  or tensor,'' \emph{Journal of Mathematics and Physics}, vol.~7, no. 1-4, pp.
  39--79, 1928.

\bibitem{TuckerDecomp}
L.~R. Tucker, ``{I}mplications of factor analysis of three-way matrices for
  measurement of change,'' in \emph{{P}roblems in measuring change}, C.~W.
  Harris, Ed.\hskip 1em plus 0.5em minus 0.4em\relax Madison WI: U of Wisconsin
  Press, 1963, pp. 122--137.

\bibitem{TTDecomp}
I.~Oseledets, ``Tensor-train decomposition,'' \emph{{SIAM Journal on Scientific
  Computing}}, vol.~33, no.~5, pp. 2295--2317, 2011.

\bibitem{PCA}
K.~Pearson, ``Liii. on lines and planes of closest fit to systems of points in
  space,'' \emph{The London, Edinburgh, and Dublin Philosophical Magazine and
  Journal of Science}, vol.~2, no.~11, pp. 559--572, 1901.

\bibitem{LPP}
X.~He and P.~Niyogi, ``Locality preserving projections,'' in \emph{Advances in
  Neural Information Processing Systems}, S.~Thrun, L.~Saul, and
  B.~Sch\"{o}lkopf, Eds., vol.~16.\hskip 1em plus 0.5em minus 0.4em\relax
  Cambridge, MA: MIT Press, 2004.

\bibitem{TensorDecomp1}
A.~Cichocki, N.~Lee, I.~Oseledets, A.-H. Phan, Q.~Zhao, and D.~P. Mandic,
  ``Tensor networks for dimensionality reduction and large-scale optimization:
  {Part 1} low-rank tensor decompositions,'' \emph{Foundations and Trends in
  Machine Learning}, vol.~9, no. 4-5, pp. 249--429, 2016.

\bibitem{TensorDecomp2}
\BIBentryALTinterwordspacing
A.~Cichocki, N.~Lee, A.~Phan, I.~Oseledets, Q.~Zhao, and D.~Mandic,
  \emph{Tensor Networks for Dimensionality Reduction and Large-Scale
  Optimization: Part 2 Applications and Future Perspectives}, ser. Foundations
  and Trends(r) in Machine Learning Series.\hskip 1em plus 0.5em minus
  0.4em\relax Now Publishers, 2017. [Online]. Available:
  \url{https://books.google.com.au/books?id=xlpkswEACAAJ}
\BIBentrySTDinterwordspacing

\bibitem{Tree-typeDecomp}
I.~Oseledets and E.~Tyrtyshnikov, ``Breaking the curse of dimensionality, or
  how to use {SVD} in many dimensions,'' \emph{SIAM Journal on Scientific
  Computing}, vol.~31, no.~5, pp. 3744--3759, 2009.

\bibitem{LPPCV}
\BIBentryALTinterwordspacing
Y.~Xu, A.~Zhong, J.~Yang, and D.~Zhang, ``Lpp solution schemes for use with
  face recognition,'' \emph{Pattern Recognition}, vol.~43, no.~12, pp.
  4165--4176, Dec. 2010. [Online]. Available:
  \url{http://dx.doi.org/10.1016/j.patcog.2010.06.016}
\BIBentrySTDinterwordspacing

\bibitem{WangAggarwalAeron2017}
W.~{Wang}, V.~{Aggarwal}, and S.~{Aeron}, ``Tensor train neighborhood
  preserving embedding,'' \emph{IEEE Transactions on Signal Processing},
  vol.~66, no.~10, pp. 2724--2732, May 2018.

\bibitem{YaleB}
A.~Georghiades, P.~Belhumeur, and D.~Kriegman, ``From few to many: Illumination
  cone models for face recognition under variable lighting and pose,''
  \emph{IEEE Transactions on Pattern Analysis and Machine Intelligence},
  vol.~23, no.~6, pp. 643--660, 2001.

\bibitem{IndianaData}
M.~F. Baumgardner, L.~L. Biehl, and D.~A. Landgrebe, ``220 band aviris
  hyperspectral image data set: June 12, 1992 indian pine test site 3,'' Sep
  2015.

\end{thebibliography}

\end{document}